%% file: main.tex
\documentclass[10pt,twocolumn,letterpaper]{article}

\usepackage[pagenumbers]{cvpr} 

\usepackage{capt-of}
\usepackage[dvipsnames]{xcolor}
\usepackage{xspace}
\usepackage{enumitem}
\usepackage{amsmath,amssymb,amsbsy,amsfonts,dsfont,pifont,bm,bbm,mathrsfs,mathtools,nicefrac}
\usepackage{algorithm,algpseudocode,listings}
\usepackage{booktabs,multirow,adjustbox,diagbox,threeparttable,tabularray}
\usepackage{bbding}
\usepackage{enumitem}

\newcommand{\tocite}[1]{{\color{red} [TO CITE]}}

\newcommand{\methodname}{MangaNinja}
\newcommand{\method}{\texttt{\methodname}\xspace}

\definecolor{cvprblue}{rgb}{0.21,0.49,0.74}
\usepackage[pagebackref,breaklinks,colorlinks]{hyperref}
\usepackage{wrapfig}
\usepackage[capitalize]{cleveref}  
\crefname{section}{Sec.}{Secs.}
\Crefname{section}{Section}{Sections}
\crefname{table}{Tab.}{Tabs.}
\Crefname{table}{Table}{Tables}
\crefname{figure}{Fig.}{Figs.}
\Crefname{figure}{Figure}{Figures}
\crefname{equation}{Eq.}{Eqs.}
\Crefname{equation}{Equation}{Equations}
\title{\methodname: Line Art Colorization with Precise Reference Following}

\author{
Zhiheng Liu$^{1,3*}$, Ka Leong Cheng$^{2,4*}$, Xi Chen$^{1,3}$, Jie Xiao$^{3}$, Hao Ouyang$^{4}$,\\
Kai Zhu$^{3}$, Yu Liu$^{3}$, Yujun Shen$^{4}$, Qifeng Chen$^{2}$, Ping Luo$^{1\dag}$\\
\\
$^1$HKU, $^2$HKUST, $^3$Tongyi Lab,$^4$Ant Group \\
}
\begin{document}

\twocolumn[{
\renewcommand\twocolumn[1][]{#1}
\maketitle
\begin{center}
    \vspace{-5pt}
    \includegraphics[width=1.0\linewidth]{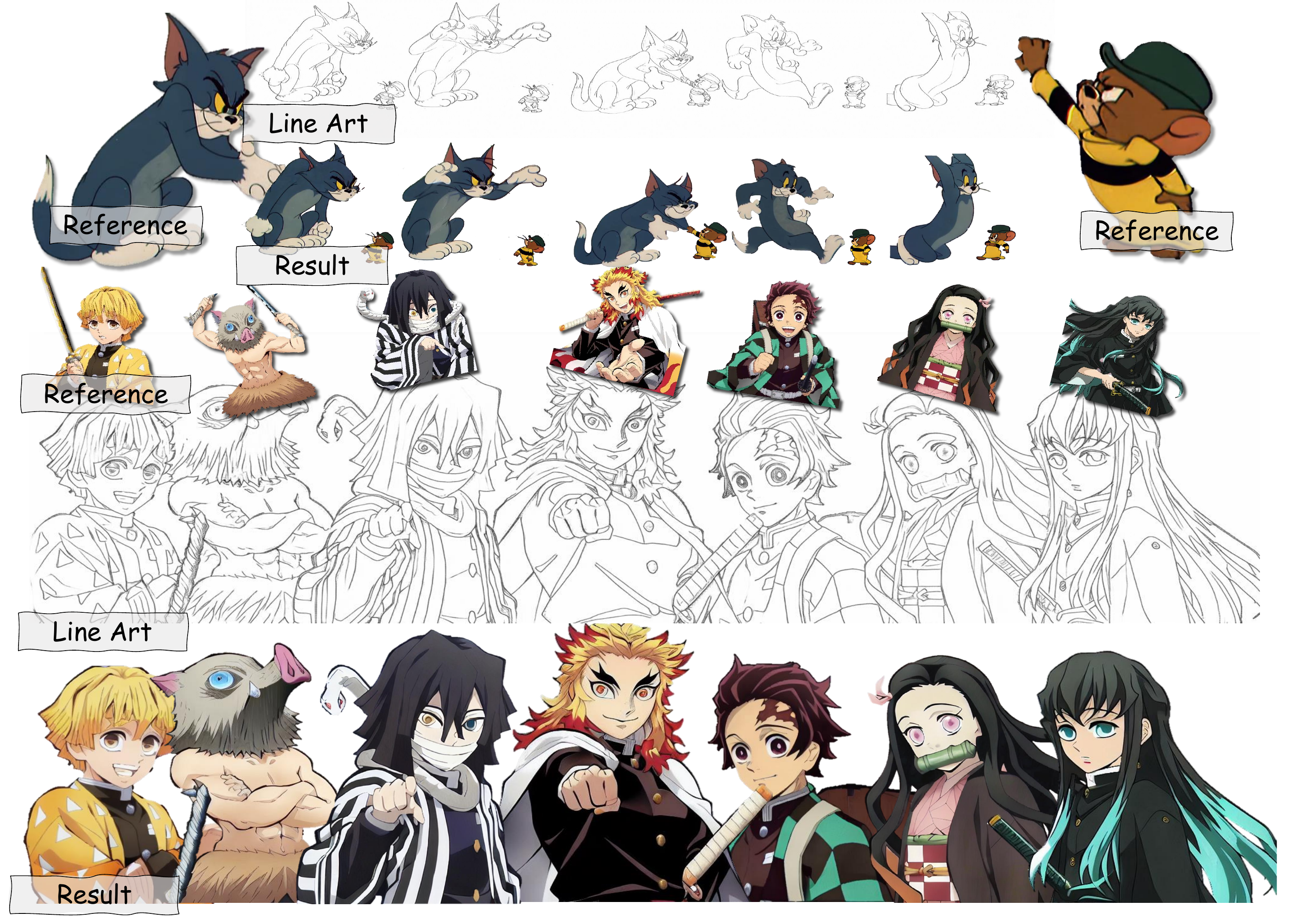}
    \captionsetup{type=figure}
    \vspace{-22pt}
    \caption{%
        \textbf{Line art colorization results.} We propose \method, a reference-based line art colorization method.
        \method automatically aligns the reference with the line art for colorization, demonstrating remarkable consistency. Additionally, users can achieve more complex tasks using point control. We hope that \method can accelerate the colorization process in the anime industry.
    }
    \label{fig:teaser}
    \vspace{8pt}
\end{center}
}]

\input{sections/0.abs}
\input{sections/1.intro}
\input{sections/2.related}
\input{sections/3.method}
\input{sections/4.exps}
\input{sections/5.conlus}
\input{sections/6.ref}

\end{document}

%% file: sections/0.abs.tex
\begin{abstract}

Derived from diffusion models, \method specializes in the task of reference-guided line art colorization.
We incorporate two thoughtful designs to ensure precise character detail transcription, including
a patch shuffling module to facilitate correspondence learning between the reference color image and the target line art,
and a point-driven control scheme to enable fine-grained color matching.
Experiments on a self-collected benchmark demonstrate the superiority of our model over current solutions in terms of precise colorization.
We further showcase the potential of the proposed interactive point control in handling challenging cases (\textit{e.g.}, extreme poses and shadows), cross-character colorization, multi-reference harmonization, \textit{etc.}, beyond the reach of existing algorithms.
%

\end{abstract}
\vspace{-20pt}

%% file: sections/1.intro.tex
\section{Introduction}
\begin{figure*}[t]
    \centering
    \includegraphics[width=\linewidth]{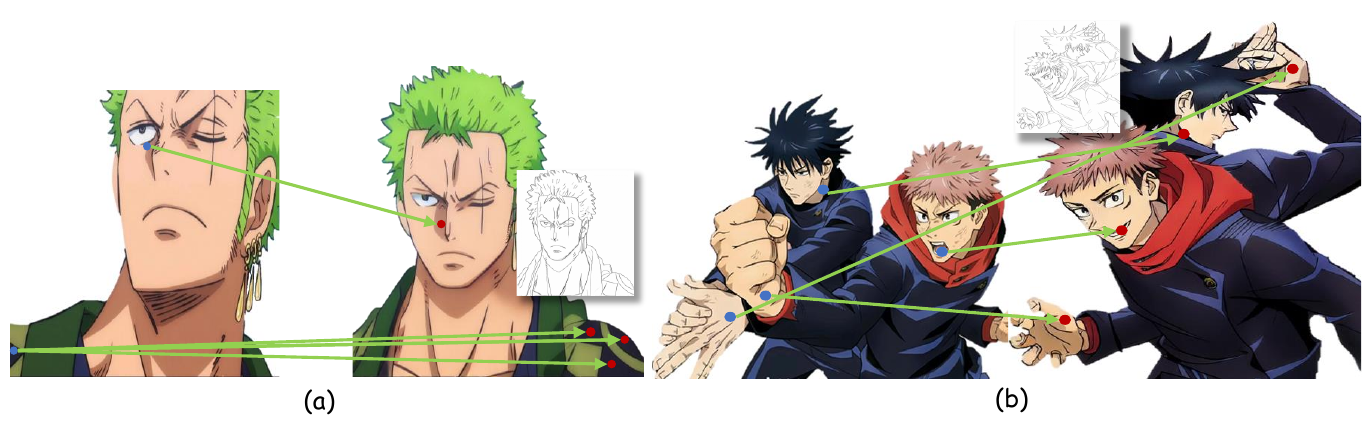}
    \vspace{-0.8cm}
    \caption{\textbf{Visualization of point guidance.} By introducing points as guidance, \method~can tackle many challenging tasks, such as when there are significant variations between reference images and line art while preserving details. See more in~\cref{e3}. 
    }
    \vspace{-0.2cm}
    \label{fig:intro}
\end{figure*}



Reference-based line art colorization aims to transform a line art image into a color image, maintaining consistency with the reference image~\cite{maejima2019graph,Zhou_2021_CVPR,li2022eliminating,carrillo2023diffusart,cao2023animediffusion}.
This technique is in high demand for comics, animation, and various other content creation applications~\cite{casey2021animation, zhang2021line,huang2022unicolor,zou2024lightweight,liang2024control}.
Unlike methods that rely solely on strokes, palettes, or text conditions~\cite{ji2022colorformer,zabari2023diffusing,utintu2024sketchdeco}, reference-based line art colorization excels in preserving both identity and semantic meaning as shown in~\cref{fig:teaser}, which is crucial for comics and manga.

Existing approaches~\cite{li2022eliminating,cao2023animediffusion} have explored reference-based colorization with fused attention mechanisms. 
However, these methods exhibit two main limitations. 
First, substantial variations between the line art and reference image often lead to semantic mismatches or confusion of colorization. 
Hence, these approaches typically demand a high standard for the reference image, requiring it to closely resemble the line art, which is impractical for real-world applications. 
Second, existing methods lack precise control, resulting in the loss of crucial details from the reference image during the colorization process.

In this paper, we introduce \method, consisting of a dual-branch structure for correspondences finding between the reference and line art images by leveraging the rich diffusion priors through cross attention.
Observing that the basic dual-branch design tends to transfer global style rather than matching local semantics, we propose a patch shuffling module, which divides the reference image into patches to encourage local matching capabilities of the model.
The patch shuffling pushes our model out of its ``comfort zone'' during optimization, facilitating it to learn an implicit matching capability that effectively handles disparities between the input line art and reference image. 

However, such semantic correspondence can still suffer from ambiguity, especially when color images include details that are hard to capture in line art (\textit{e.g.}, nose shading in \cref{fig:intro}a), when some elements in the line art occupy only a small area of the reference image (\textit{e.g.}, shoulder garment pattern in \cref{fig:intro}a), or when significant variations and complex compositions create semantic confusion (\textit{e.g.}, multiple characters in \cref{fig:intro}b). 
To further support finer-grained coloring matching, we introduce a point-driven control scheme powered by PointNet, which offers detailed control using user-defined cues in an interactive manner.
During experiments, we find that point control only works when the model is aware of local semantics, highlighting the importance and effectiveness of patch shuffling.


We take advantage of the inherently natural semantic correspondences and visual variances presented in anime videos to construct training data pairs.
Specifically, we randomly select two frames from a video: one serves as the reference for the Reference U-Net, while the other, along with its line art version, acts as the target and input for the Denoising U-Net.
As for the explicit correspondence, we employ an off-the-shelf model to label matching points in the training image pairs, encode these points with PointNet, and integrate them into the main branch via attention. 
With our carefully designed patch shuffling strategy and point-driven control scheme, \method~effectively manages challenging scenarios, such as varying poses or details missing between reference and line art, multi-reference inputs, and colorization with discrepant references, as shown in \cref{e3}. 
It excels in complex colorization tasks, producing high-quality results from line art while accurately preserving character identity, as demonstrated in \cref{fig:teaser}.
For a fair and systematic evaluation, we construct a comprehensive benchmark for line art colorization. 
Our extensive quantitative and qualitative experiments demonstrate that our approach outperforms existing baselines, achieving state-of-the-art results in visual fidelity and identity preservation, making it beneficial for comics, animation, and various content creation applications.

%% file: sections/2.related.tex
\section{Related Work}

\subsection{Line Art Colorization}

Line art colorization aims to fill the blank regions of line art with appropriate colors. Currently, several user-guided colorization techniques exist, including text prompts~\cite{zhang2023adding,cao2021line,kim2019tag2pix}, scribble~\cite{dou2021dual,cao2021line,zhang2021user,zhang2018two,liu2018auto,sangkloy2017scribbler}, and reference image~\cite{InclusionMatching2024,wu2023flexicon,wu2023self,li2022eliminating,li2022style,zhang2021line}. 
However, text-based and scribble methods have limitations in achieving precise color filling for the overall line art. 
Existing reference-based colorization approaches often have limited performance due to inaccurate structural and semantic matching, particularly when there are substantial differences between the reference image and the line art. 
Moreover, in practical applications, more complex scenarios arise, such as requiring multiple reference images to handle the colorization of various elements in the line art.
Consequently, it is challenging to seamlessly integrate the existing line art colorization methods into the animation industry workflow.
Our approach leverages priors from pretrained diffusion models and enhances the model's matching capabilities by learning from video data, allowing users to accomplish complex colorization tasks with simple point guidance.

\subsection{Visual Correspondence}
In computer vision, correspondence~\cite{zabih1994non} involves identifying and matching related features or points across different images, often used for tasks such as stereo vision~\cite{agarwal2011building,ozyecsil2017survey,schonberger2016structure,schonberger2016pixelwise}, motion tracking~\cite{gao2022aiatrack,yan2022towards}.
Traditional methods use hand-crafted features~\cite{lowe2004distinctive,bay2006surf} to find correspondences, whereas recent deep learning approaches~\cite{cho2021cats,lee2021patchmatch,kim2022transformatcher,huang2022learning} leverage supervised learning with labeled data to learn matching capabilities.
However, due to the requirement for precise pixel-level annotations, these methods struggle to scale up, as such detailed labeling is challenging and expensive.
Later, researchers begin exploring the establishment of weakly supervised~\cite{wang2020learning} or self-supervised~\cite{wang2019learning,jabri2020space} visual correspondence models.
Recent studies \cite{tang2023emergent,hedlin2024unsupervised,peebles2022gan} show that the rich priors inherent in the latent representations of generative pretrained models like GAN~\cite{goodfellow2020generative} and Diffusion~\cite{sohl2015deep} models can be utilized to identify visual correspondence.
Leveraging the inherent rich priors of correspondences in pre-trained diffusion models, our method achieves reference-based colorization by learning to match between line art and reference images.

\subsection{Diffusion-based Consistent Generation}
Consistent generation based on pretrained diffusion models can be categorized into three main directions. 
The first direction leverages a training-free or rapid fine-tuning strategy for image editing~\cite{shi2024dragdiffusion, ling2023freedrag,brooks2023instructpix2pix,kawar2023imagic,cao2023masactrl,mao2023guided,tumanyan2023plug,mou2023dragondiffusion,bar2022text2live,hertz2022prompt,liew2022magicmix}, where they conduct global or local editing by modifying text prompts or introducing new guidance to adjust the attention layers.
However, they generally struggle with robustness in challenging scenarios and rely heavily on the input guidance signals.
The second direction is customized generation~\cite{safaee2024clic,gu2024mix,tang2024realfill,ruiz2023dreambooth,kumari2023multi,liu2023cones2,liu2023cones,avrahami2023break,gal2022image}, which generally involves fine-tuning on 3 to 5 example images per concept, where some methods may take about half an hour for a single concept.
The third direction involves further training the pretrained diffusion model with extensive domain-specific data, learning to incorporate encoded image features into the main denoising network~\cite{pan2024locate,yuan2023customnet,zhang2023controlcom,xie2023dreaminpainter}.
For instance, Paint-by-Example~\cite{yang2022paint} and ObjectStitch~\cite{song2023objectstitch} utilize CLIP~\cite{radford2021learning} to encode images for extracting object representations, while AnyDoor~\cite{chen2023anydoor} collects training samples from videos and employs the DINOv2~\cite{oquab2023dinov2} as the image encoder.
However, these methods primarily focus on general objects in images, lacking fine-grained matching capabilities. 

%% file: sections/3.method.tex
\section{Method}
\begin{figure*}[t]
    \centering
    \includegraphics[width=0.96\linewidth]{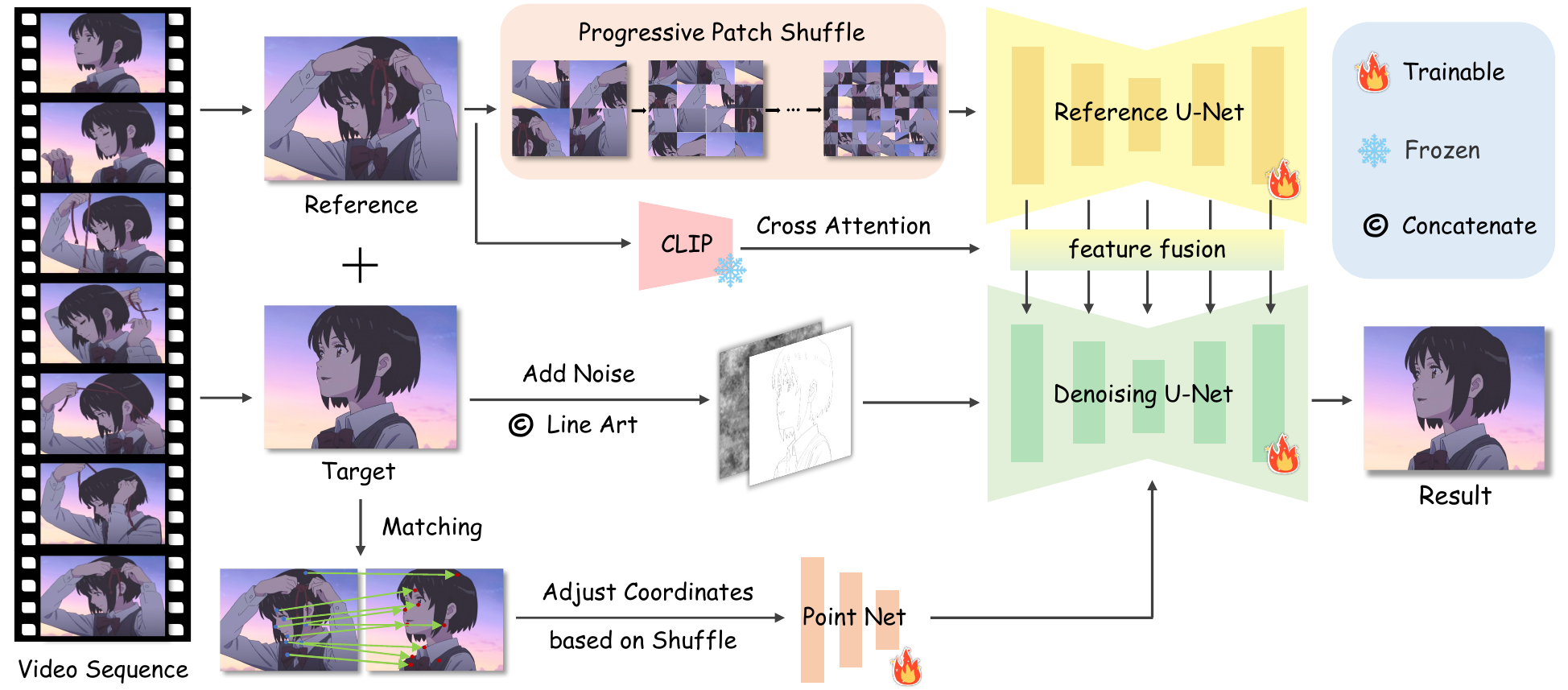}
    \caption{\textbf{The training process of \method.} We randomly select two frames from video data, using one frame as a reference image and extracting the line art from the other. 
    Both frames are input into the Reference U-Net and the Denoising U-Net, respectively. 
    To enhance the model's automatic matching and fine-grained control capabilities, we propose a series of training strategies, including progressive patch shuffling.
    Additionally, we employ an off-the-shelf model to extract matching points from the two frames, and these point maps are fed into the main branch through PointNet. 
    }
    \label{fig:pipeline}
    \vspace{-0.35cm}
\end{figure*}
%
%
%

\subsection{Overall Pipeline}\label{m1}
The overall framework of \method~is presented in~\cref{fig:pipeline}. 
Our goal is to match and colorize, producing a vibrant anime image $I_{\rm target}$ from a line art $I_{\rm line}$ and a reference image $I_{\rm ref}$ of the same character. 
Additionally, users can pre-define specific points $P_{\rm ref}$ on the reference image and their corresponding points $P_{\rm line}$ on the line art.
Guided by the matching points, the model ensures color consistency during the colorization process, thereby achieving fine-grained control and excellent performance even in challenging scenarios.

Anime video sequences inherently present identity consistency across frames while simultaneously exhibiting various spatial and temporal transformations. 
These transformations include, but are not limited to, scale variations (e.g., zooming effects), changes in object orientation, and alterations in pose.
Thanks to such property, we construct training image pairs by randomly sampling two distinct frames from a video clip.
The first frame serves as the reference, and we employ an off-the-shelf line art extraction model~\cite{zhang2023adding} to derive the line art from the second frame, which serves as the target image.
During training, we use LightGlue~\cite{lindenberger2023lightglue}, a state-of-the-art point-matching algorithm, to extract corresponding point pairs between two frames.
%

%

%
\subsection{Architecture Design}\label{m2}
\noindent\textbf{Reference U-Net.}
Given the stringent detail requirements in line art colorization, the main challenge is how to effectively encode the reference image for finer-grained feature extraction.
Recent studies~\cite{hu2023animateanyone,xu2023magicanimate} demonstrate the effectiveness of leveraging an additional U-Net architecture to address this issue, and we are inspired to introduce a Reference U-Net using a similar design.
After encoding the reference image into a 4-channel latent representation using VAE, it is fed into the Reference U-Net to extract multi-level features for fusion with the main Denoising U-Net.
Specifically, we concatenate the key and value from the self-attention layers of both the reference and denoising branches, as described in~\cref{eq:concat}, injecting the multi-level reference features into the corresponding layers of the Denoising U-Net.
\begin{equation}
    {\rm Attn}= {\rm softmax} ( \frac{Q_{\rm tar} \left[K_{\rm tar}, K_{\rm ref}\right]^{\top}}{\sqrt{d}} ) [V_{\rm tar}, V_{\rm ref}].
    \label{eq:concat}
\end{equation}

\noindent\textbf{Denoising U-Net.}
The main branch utilizes the Reference U-Net and PointNet as conditions for image colorization. We extract the line art from the images using LineartAnimeDetector~\cite{zhang2023adding}, then replicate the single-channel line art three times to input into the variational autoencoder (VAE) for compression into the latent space. Next, we concatenate this with the noisy image latent, resulting in a total of 8 channels. Additionally, we experiment with sending the line art through ControlNet~\cite{zhang2023adding} and find that both approaches yield comparable performance. For resource efficiency, we opt for the first method. Furthermore, we replace the original text embeddings with image embeddings extracted from the CLIP encoder.

\noindent\textbf{Progressive patch shuffle for local matching.}
Although we inject the reference image features layer by layer into the Denoising U-Net, we observe that the strong structural cues provided by the line art enable easy coarse global matching, which hinders the learning of detailed matching ability.
To address this, we propose a progressive patch shuffle strategy. Specifically, we divide the reference image into multiple small patches and randomly shuffle them to disrupt the overall structural coherence, as shown in~\cref{fig:pipeline}.
The idea behind this technique is to encourage the model to focus more on smaller patches (even at the pixel level) within the reference image to achieve finer-grained, local matching abilities rather than global ones.
Moreover, we adopt a coarse-to-fine learning scheme by progressively increasing the number of randomly shuffled patches from $2 \times 2$ to $32 \times 32$.
Apart from the shuffling technique, we also employ some common data augmentation techniques, such as random flipping and rotation, to increase the variation between the reference and target image. 

\subsection{Fine-grained Point Control}\label{m3}
However, such semantic correspondence can still suffer from ambiguity, especially when color images contain details that are difficult to capture in line art. Moreover, users often require a simple interactive method to handle complex tasks. To address this, we design a point-based fine-grained control mechanism and propose a series of strategies to enhance the effectiveness of point control.

\noindent\textbf{Point embedding injection.}
We represent user-specified matching point pairs using two point maps, each being a single-channel matrix matching the input image's resolution.
For each matching point pair, we assign the same unique integer values to their respective coordinates on both point maps, with all other positions set to $0$.
During training, we randomly select up to $24$ matching point pairs, with the option to select zero points as well.
Hence, users can opt not to indicate matching points for control during inference, instead fully relying on the autonomous matching capability of the model.

We propose a PointNet composed of multiple convolutional layers and SiLU activation functions to encode the point maps as multi-scale embeddings. Similarly, the point embeddings $E_{\rm tar}$ and $E_{\rm ref}$ are integrated into the main branch via a cross-attention mechanism by adding them to the query and key, as described in~\cref{eq:point}:
\begin{equation}
    {\rm Attn}= {\rm softmax} ( \frac{Q'_{\rm tar}[K'_{\rm tar}, K'_{\rm ref}]^{\top}}{\sqrt{d}} ) [V_{\rm tar}, V_{\rm ref}],
    \label{eq:point}
\end{equation}
where $Q'_{\rm tar} = Q_{\rm tar} + E_{\rm tar}$, $K'_{\rm tar} = K_{\rm tar} + E_{\rm tar}$, and $K'_{\rm ref} = K_{\rm ref} + E_{\rm ref}$.
%

\noindent\textbf{Multi classifier-free guidance.}
To individually control the guiding strength of the reference image and the points during the generation inference process, we employ multiple classifier-free guidance:
\begin{equation}
\begin{aligned}
\epsilon_\theta&(z_{\rm t}, c_{\rm ref}, c_{\rm points}) = \epsilon_\theta(z_{\rm t}, \emptyset, \emptyset) \\
&\quad + \omega_{ref} \big( \epsilon_\theta(z_{\rm t}, c_{\rm ref}, \emptyset) - \epsilon_\theta(z_{\rm t}, \emptyset, \emptyset) \big) \\
&\quad + \omega_{\rm points} \big( \epsilon_\theta(z_{\rm t}, c_{\rm ref}, c_{\rm points}) - \epsilon_\theta(z_{\rm t}, c_{\rm ref}, \emptyset) \big),
\end{aligned}
\end{equation}
where \( c_{\rm ref} \) denotes the condition input from the reference image via the Reference U-Net, while \( c_{\rm points} \) denotes the condition input from the user-specified points through the PointNet. 
Increasing \( \omega_{\rm ref} \) makes the model rely more on its automatic matching capabilities. However, when we want to use points as guidance to accomplish more complex tasks (see~\cref{e3}), we should increase \( w_{\rm points} \) to amplify the influence of the points.

\noindent\textbf{Condition dropping.}
To enhance the model's reliance on sparse point-based control signals, we randomly drop the line art condition during training.
Without the structural guidance of the line art, we prompt the model to reconstruct the target image $I_{\rm target}$ from the reference image $I_{\rm ref}$, relying solely on the sparse yet precise matching indicated by the point pairs $P_{\rm ref}$ and $P_{\rm line}$.
This helps compel our model to learn the precise point-based control more effectively.


%
%
%


%
%

\noindent\textbf{Two-stage training.}
To further amplify the effects of precise point-based control, we design a two-stage training strategy. 
In the first stage, we adopt condition dropping for both the reference image and point signals for unconditional generation, where the model concurrently learns the abilities to extract corresponding reference features and leverage the specified point correspondences for colorization.
In the second stage, we only train the PointNet module, thereby enhancing the ability of PointNet to encode point maps, leading to stronger point control.
%


%
%
%
%

\begin{figure*}[t]
    \centering
    \includegraphics[width=1.0\linewidth]{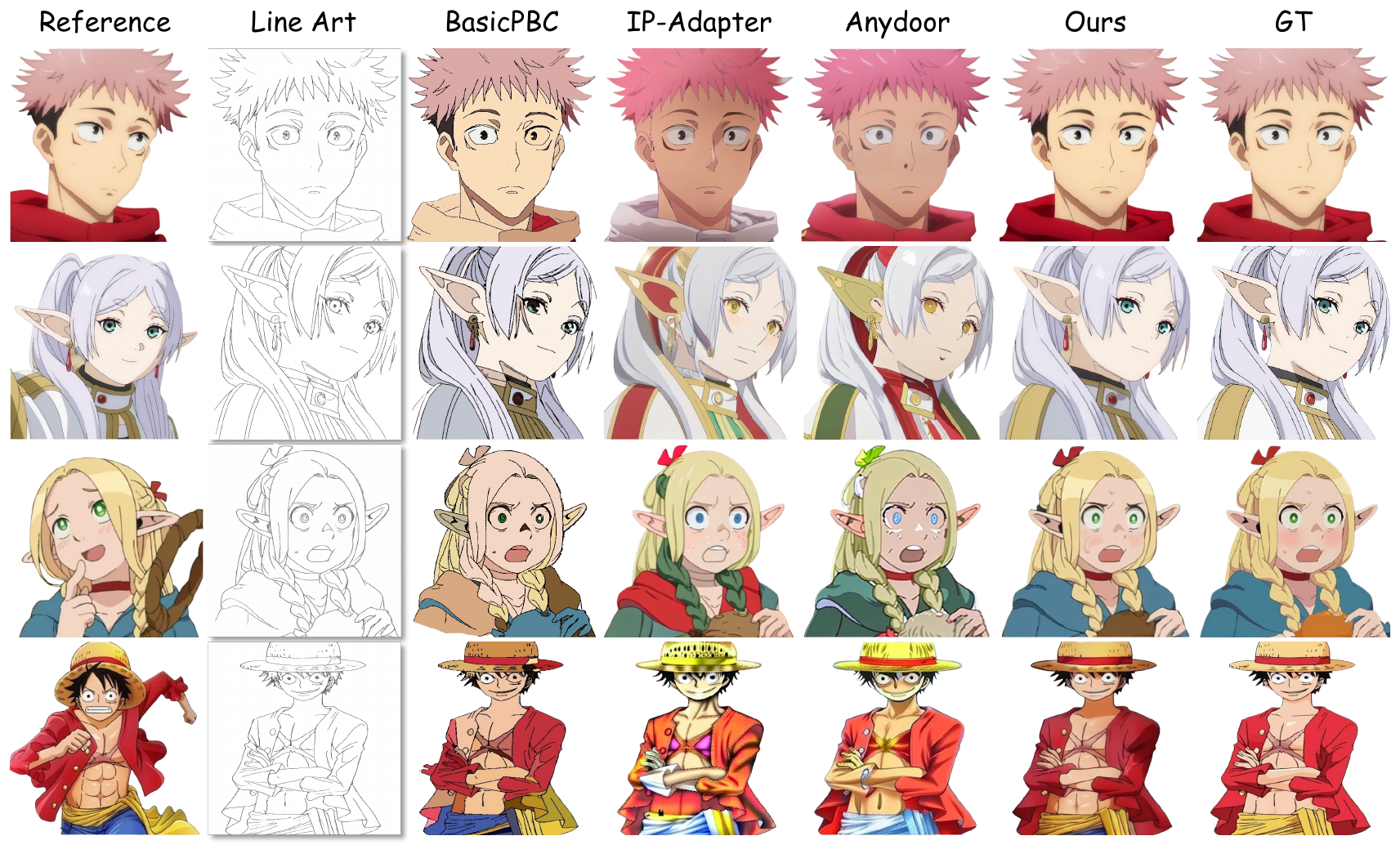}
    \caption{\textbf{Qualitative comparisons.} We compare our method with the state-of-the-art non-generative colorization method BasicPBC, the consistency generation method IP-Adapter, and AnyDoor. The results demonstrate that our method significantly outperforms them in terms of colorization accuracy and generated image quality. Notably, our method does not use points for guidance in the generated results.}
    \label{fig:comparison}
\end{figure*}

\subsection{Evaluation Benchmark}\label{m4}
%
Existing works such as BasicPBC~\cite{InclusionMatching2024} and Animediffusion~\cite{cao2023animediffusion} design test sets that focus only on specific domains, involve minimal discrepancy between the reference and target images, and evaluate using inconsistent metrics. 
%
%
Therefore, we see the crucial need to establish a comprehensive and consistent evaluation benchmark.
We construct a benchmark to systematically evaluate the
performance of line art colorization. 
Specifically, we collect $200$ image pairs of the same character from various anime, encompassing both human and non-human characters with diverse facial expressions and appearances, including attire.
Each evaluation sample consists of a target image with its line art estimated using an off-the-shell LineartAnimeDetector model~\cite{zhang2023adding} and a reference image as colorization guidance.
In the colorization process, the focus is typically on the foreground character portions, so we segment all images to extract the foreground subjects.
Moreover, we follow the methodology outlined in DreamBooth~\cite{ruiz2023dreambooth} to calculate the CLIP~\cite{radford2021learning} and DINO~\cite{oquab2023dinov2} semantic image similarities between the generated images and the ground truth.
Furthermore, to assess the quality of the generated images, we compute the Peak Signal-to-Noise Ratio (PSNR) and the Multi-Scale Structural Similarity Index (MS-SSIM)~\cite{wang2004image}.
Meanwhile, to evaluate coloring accuracy in complex tasks—such as those involving multiple references or colorization with differing reference points mentioned in~\cref{e3}—we require a more granular evaluation at the pixel level.
Specifically, we annotate $50$ predefined pairs of matching points for each image pair; for evaluation we employ the mean squared error (MSE) for the $3 \times 3$ patches centered around each pair of matching points.

%% file: sections/4.exps.tex
\section{Experiments}
\subsection{Implementation Details}\label{e1}
\noindent\textbf{Training details.}
For training \method~, we utilize a dataset, sakuga-42m~\cite{pan2024sakuga}, which comprises 42 million keyframes covering a wide range of artistic styles, geographical regions, and historical periods.
We eliminate excessively similar duplicate frames by calculating the Structural Similarity Index (SSIM). Furthermore, we set the frame interval between the reference and target frames to 36, excluding videos that are too short. Ultimately, we retain 300,000 video clips.
We initialize both the Reference and Denoising U-Net with pre-trained weights sourced from Stable Diffusion 1.5~\cite{rombach2022high}
The training process spans 200k steps (with the first stage lasting 180k steps and the second stage 20k steps), starting with an initial learning rate of $10^{-3}$, which decays every 30k steps. The entire training process is completed within one day using eight A100-80G GPUs.
\subsection{Comparisons}\label{e2}
In this section, we compare with the current state-of-the-art line art colorization method, BasicPBC~\cite{InclusionMatching2024}. Additionally, we also conduct comparisons with several generative methods that can achieve similar functions.
These include IP-Adapter~\cite{ye2023ip-adapter}, which serves as an adapter to enhance the image prompting capabilities of pretrained text-to-image diffusion models, and Anydoor~\cite{chen2023anydoor}, a zero-shot object-level image customization method.
In addition, we will discuss the cartoon interpolation method ToonCrafter~\cite{xing2024tooncrafter} in the supplementary materials, as the official repository has not yet released its colorization function and exhibits poor performance when there are significant discrepancies.

\noindent\textbf{Qualitative comparison.}
We visualize the comparison results in~\cref{fig:comparison}. BasicPBC samples colors in the vicinity of the corresponding area in the line art; hence, the generated results can be unsatisfactory when there is a large discrepancy between the reference and the line art.
Moreover, as the model itself does not have a generative capability, it does not perform well in handling light and shadow.
For generative methods, we introduce a controlnet for the IP-Adapter and AnyDoor, and carefully annotate the masks of the reference region, then feed them to AnyDoor.
Leveraging the strong prior knowledge of pre-trained models, the generated results become more natural. Compared to the IP-Adapter, AnyDoor better retains the color details of the reference image. However, neither method possesses fine-grained matching capability and can only achieve coarse colorization results, leading to serious color confusion.
Notably, our method does not use points for guidance in the generated results. This is because, during the training process, our method learns from image pairs in video data and gradually shuffles the reference image at the patch level from simple to complex, which endows the model with excellent matching capability. Benefiting from the design of the point, as shown in~\cref{e3}, our method also excels in some more complex scenarios.

\noindent\textbf{Quantitative comparison.}
We conduct a quantitative comparison using our constructed benchmark. It is worth noting that this benchmark contains 200 pairs of images, which means we perform a total of 400 inferences (interchanging the reference image and ground truth). The results are presented in~\cref{tab:comparison_methods}.
The results indicate that the BasicPBC outperforms generative methods in pixel-level evaluation metrics. However, it is noteworthy that BasicPBC performs weaker in terms of image feature similarity metrics compared to these methods. Additionally, Anydoor requires manual labeling of masks in reference images to achieve better performance. In contrast, our approach demonstrates a significant advantage over previous methods in both pixel-level and image feature similarity metrics.


\begin{table}[t]
    \centering
    \small
    \caption{
        \textbf{Quantitative comparison.} \method~demonstrates superior performance across both objective and perceptual metrics. AnyDoor: without mask; AnyDoor*: with mask. Ours: no point guidance; Ours (full): with point guidance.
    }
    \label{tab:comparison_methods}
    \vspace{-8pt}
    \SetTblrInner{rowsep=1.2pt}      
    \SetTblrInner{colsep=1.6pt}      
    \begin{tblr}{
        cells={halign=c,valign=m},   
        column{1}={halign=l},        
        hline{1,2,6,8}={1-6}{},       
        hline{1,8}={1.0pt},          
        vline{2}={1-7}{},         
    }
    Method                            & DINO $\uparrow$ & CLIP $\uparrow$ &  PSNR $\uparrow$ & MS-SSIM $\uparrow$ & LPIPS $\downarrow$ \\
    BasicPBC~\cite{InclusionMatching2024}                  & 42.64               & 79.64              & 17.58           & 0.894              & 0.33  \\
    IP-Adapter~\cite{ye2023ip-adapter}                & 55.42               & 82.39              & 16.19           & 0.845             & 0.30  \\
    Anydoor~\cite{chen2023anydoor} & 51.36               & 80.73              & 15.12           & 0.827              & 0.32  \\
    AnyDoor*~\cite{chen2023anydoor}  & 63.79               & 83.91              & 16.24           & 0.874              & 0.27  \\
    Ours                              & \underline{68.23}      & \underline{88.34}     & \underline{20.37}  & \underline{0.962}     & \underline{0.22} \\
    Ours (full) & \textbf{69.91}  & \textbf{90.02} & \textbf{21.34} & \textbf{0.972} & \textbf{0.21}
    \end{tblr}
    \vspace{-5pt}
\end{table}

\begin{table*}[t]
    \centering
    \small
    \caption{
        \textbf{Ablation study} on the effect of various training strategies. The first five evaluation metrics assess the overall quality of the coloring results, while the MSE metric evaluates the coloring accuracy at the specified matching pixels. The base model indicates training solely with video data, without employing any strategies; the full model incorporates all training strategies. Gray numbers in parentheses represent statistics assessed without point guidance.
    }
    \label{tab:asb}
    \vspace{-8pt}
    \SetTblrInner{rowsep=1.2pt}      
    \SetTblrInner{colsep=5.2pt}      
    \begin{tblr}{
        cells={halign=c,valign=m},   
        column{1}={halign=l},        
        hline{1,2,7,8}={1-7}{},       
        hline{1,8}={1.0pt},          
        vline{2}={1-7}{},         
    }
    \                                                                              & DINO Sim $\uparrow$                      & CLIP Sim $\uparrow$                      & PSNR $\uparrow$                          & MS-SSIM $\uparrow$                      & LPIPS $\downarrow$                     & MSE $\downarrow$ \\
    \uppercase\expandafter{\romannumeral1}. base model                             & 64.13 \textcolor{gray}{(63.91)}          & 85.05 \textcolor{gray}{(84.75)}          & 18.12 \textcolor{gray}{(18.02)}          & 0.914 \textcolor{gray}{(0.912)}         & 0.26 \textcolor{gray}{(0.27)}          & 0.0151           \\
    \uppercase\expandafter{\romannumeral2}. base model + condition dropping        & 64.92 \textcolor{gray}{(64.79)}          & 85.44 \textcolor{gray}{(85.22)}          & 19.02 \textcolor{gray}{(18.61)}          & 0.941 \textcolor{gray}{(0.929)}         & 0.25 \textcolor{gray}{(0.25)}          & 0.0125           \\
    \uppercase\expandafter{\romannumeral3}. base model + progressive patch shuffle & 67.78 \textcolor{gray}{(67.12)}          & 87.42 \textcolor{gray}{(86.93)}          & 20.18 \textcolor{gray}{(19.72)}          & 0.956 \textcolor{gray}{(0.952)}         & 0.23 \textcolor{gray}{(0.23)}          & 0.0091           \\
    \uppercase\expandafter{\romannumeral4}. base model + multi cfg                 & 64.63                                    & 86.02                                    & 18.74                                    & 0.943                                   & 0.24                                   & 0.0133           \\
    \uppercase\expandafter{\romannumeral5}. base model + two-stage training        & 64.32                                    & 86.34                                    & 19.36                                    & 0.939                                   & 0.24                                   & 0.0113           \\
    \uppercase\expandafter{\romannumeral6}. full model                             & \textbf{69.91 \textcolor{gray}{(68.23)}} & \textbf{90.02 \textcolor{gray}{(88.34)}} & \textbf{21.34 \textcolor{gray}{(20.37)}} & \textbf{0.972 \textcolor{gray}{(0.962)}} & \textbf{0.21 \textcolor{gray}{(0.22)}} & \textbf{0.0072}  \\
    \end{tblr}
    \vspace{-5pt}
\end{table*}

\subsection{Challenging Cases with Point Guidance}\label{e3}
\begin{figure}[t]
    \centering
    \includegraphics[width=0.9\linewidth]{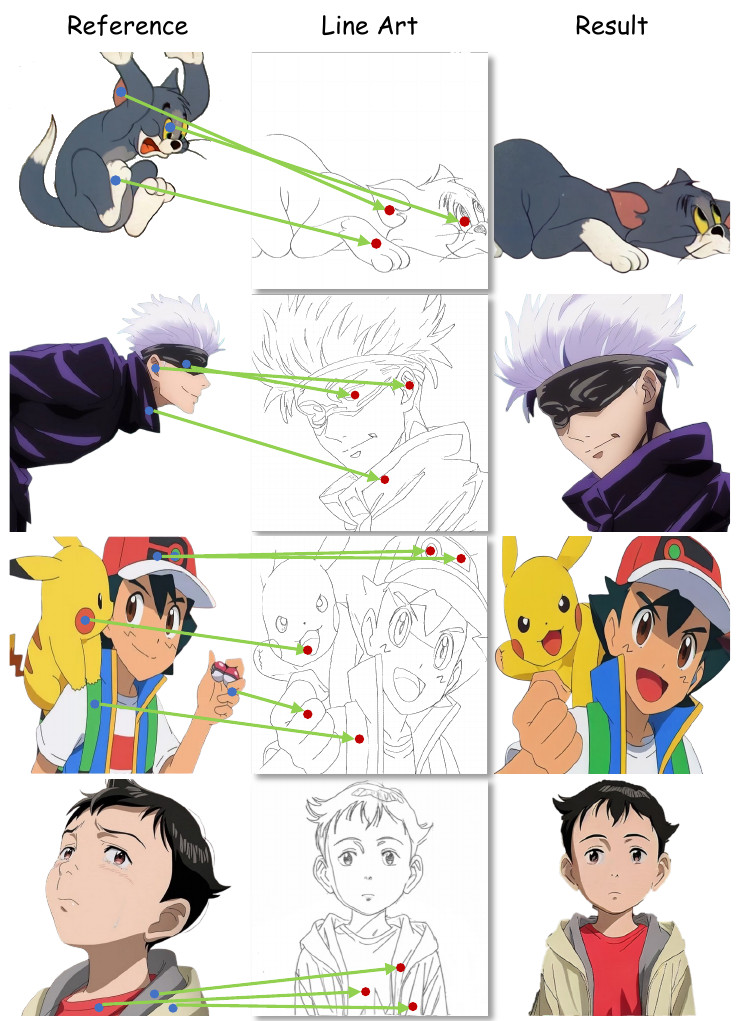}
    \caption{\textbf{Visualization of varying poses or missing details.} With point guidance, \method~can tackle many challenging cases. For instance, in the first two rows, there are significant variations between the reference image and line art. 
    %
    Furthermore, users can employ point guidance to colorize regions or elements with no matches in the reference; for example, the lower parts of the clothing are missing in the reference image of the third sample.
    When dealing with multiple objects, point guidance effectively prevents color confusion, as demonstrated in the last row.}
    \label{fig:hard}
    \vspace{-0.6cm}
\end{figure}

\noindent\textbf{Varying poses or missing details.}
As shown in~\cref{fig:hard}, we present some more challenging examples of line art colorization.
As demonstrated in the first row, even with substantial variations between the line art and the reference image, excellent colorization can be achieved with points serving as guidance.
Furthermore, the reference image sometimes lacks certain elements present in the line art, as exemplified in the first column of the second row.
The line art includes the complete garment, but the reference image only provides the upper half. 
With \method~, users can color the lower half of the clothes guided by points, utilizing the upper half from the reference image.
Finally, as shown in the second column of the second row, there may be multiple objects in the line art that interact with each other. 
Segmenting the line art and coloring each part separately can sometimes result in inaccuracies and additional costs. 
However, with the guidance of points, \method~can achieve one-time colorization of multiple objects with good performance.

\noindent\textbf{Multi-ref colorization.}
\begin{figure}[t]
    \centering
    \includegraphics[width=1.0\linewidth]{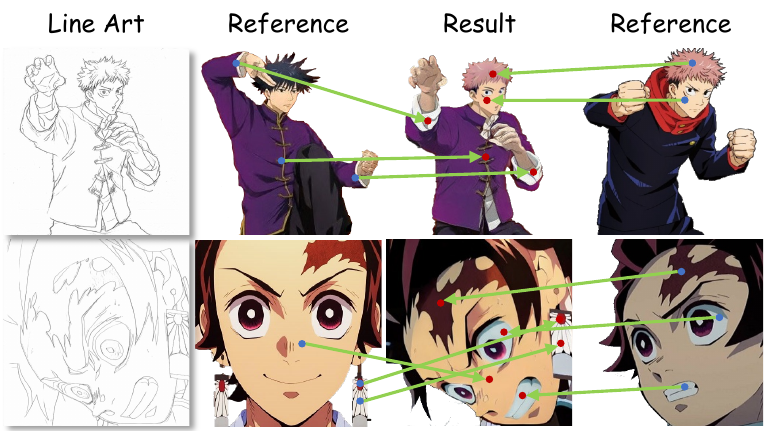}
    \caption{\textbf{Visualization of multi-ref colorization.} \method~enables users to select specific areas from multiple reference images through points, providing guidance for all elements in the line art. 
    Additionally, it effectively resolves conflicts between similar visual elements across the reference images.}
    \label{fig:multi}
    \vspace{-0.45cm}
\end{figure}
As demonstrated in~\cref{fig:multi}, in practical applications, a single reference image may not always encompass all the elements in line art that require colorization. 
Benefiting from the point-guided design, our method allows for the simultaneous use of multiple reference images for colorization. 
Specifically, Users can combine multiple images and input them into Reference U-Net, which then employs points to match different regions from the reference images with corresponding elements in the line art. This approach facilitates many-to-one colorization and effectively resolves content conflicts among the various reference images.

\begin{figure}[t]
    \centering
    \includegraphics[width=0.9\linewidth]{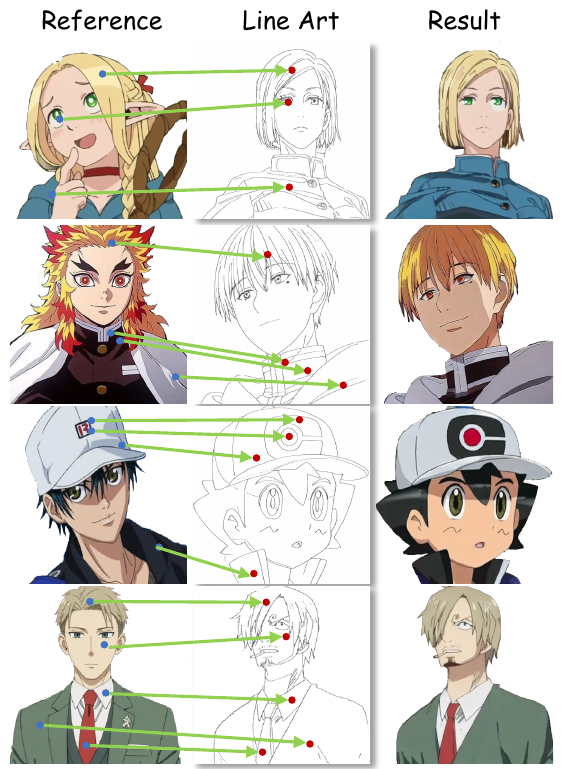}
    \caption{\textbf{Visualization of colorization with discrepant reference.} Our method uses points as guidance to achieve semantic color matching with fine control. 
    We believe this interactive colorization with discrepant references can effectively assist users in their colorization attempts and inspire new ideas. }
    \label{fig:discrepant}
    \vspace{-0.45cm}
\end{figure}

\noindent\textbf{Colorization with references of different characters.}
\method~is trained on a large number of image pairs from video data, which provides it with semantic matching capability and excellent generalization properties. 
Moreover, by utilizing point guidance, we can achieve precise colorization. Consequently, even when the reference image and the line art are different characters, the model can still perform colorization effectively.
As illustrated in~\cref{fig:discrepant}, users can take advantage of this capability and engage in an interactive process to explore and find inspiration for colorization.
\subsection{Ablation Studies}\label{e4}

\noindent\textbf{Ablation of training strategies.}
We conduct a series of ablation studies in~\cref{tab:asb} to investigate how different training strategies impact the colorization performance and matching capability of our model.
The first five metrics assess the overall quality of the colorization, while the MSE measures the accuracy of the color predictions at the pixel locations of the guiding points.
The ablation performance of our model with point guidance is shown in black; to further demonstrate the model's ability for automatic color matching, we present the ablation results as gray numbers in parentheses, representing evaluations done without point guidance.

The ablation experiments demonstrate that all strategies contribute to improving point-guided generation, enabling our method to address a broader range of complex tasks. 
Notably, even without using points as guidance, both condition dropping and progressive patch shuffle enhance the model's automatic matching capability, with the latter yielding the most notable improvement. 
Specifically, it disrupts the reference image's inherent structural patterns during training, enabling the model to learn local matching capabilities.
Only after learning this local matching ability does the effect of point guidance become clearly evident. 
Meanwhile, we provide a further analysis of the progressive patch shuffle in the supplementary materials.
%
%
%
%
%

%% file: sections/5.conlus.tex
\section{Conclusion}

In this work, we present~\method~, a novel reference-guided line art colorization method. 
Through a series of training strategies, our method utilizes a dual-branch structure and PointNet to achieve precise automatic matching while also allowing users to exert fine-grained control by defining matching points. 
\method~exhibits impressive performance in complex scenarios, including discrepant reference colorization, significant variations between reference images and line art, and multi-subject colorization. 
Additionally, we propose a benchmark for the standardized evaluation of reference-based colorization. Our work serves as a practical tool to accelerate the coloring process in the anime industry while inspiring future research in colorization.

%% file: sections/6.ref.tex
\bibliographystyle{ieeenat_fullname}
\bibliography{main}